\documentclass[runningheads]{llncs}

\usepackage{eccv}

\usepackage{eccvabbrv}

\usepackage{graphicx}
\usepackage{booktabs}
\usepackage[accsupp]{axessibility}  % Improves PDF readability for those with disabilities.

\usepackage{tablefootnote}

\usepackage{hyperref}

\usepackage{orcidlink}

\usepackage{soul}

\begin{document}

% ---------------------------------------------------------------
% TODO REVIEW: Replace with your title
\title{Modelling the Distribution of Human Motion for Sign Language Assessment} 

% TODO REVIEW: If the paper title is too long for the running head, you can set
% an abbreviated paper title here. If not, comment out.
% \titlerunning{Abbreviated paper title}

% TODO FINAL: Replace with your author list. 
% Include the authors' OCRID for the camera-ready version, if at all possible.
\author{Oliver Cory\inst{1}\orcidlink{0009-0002-5383-7202} \and
Ozge Mercanoglu Sincan\inst{1}\orcidlink{0000-0003-3285-8020} \and
Matthew Vowels\inst{1,2,3}\orcidlink{0000-0002-8811-1156} \and
Alessia Battisti\inst{4}\orcidlink{0000-0002-1696-6921} \and
Franz Holzknecht\inst{5}\orcidlink{0000-0002-1218-2062} \and
Katja Tissi\inst{5} \and
Sandra Sidler-Miserez\inst{5} \and
Tobias Haug\inst{5}\orcidlink{0000-0002-8713-1163} \and
Sarah Ebling\inst{4}\orcidlink{0000-0001-6511-5085} \and
Richard Bowden\inst{1}\orcidlink{0000-0003-3285-8020}}

\authorrunning{O. Cory et al.}

\institute{Centre for Vision, Speech and Signal Processing, University of Surrey, Guildford, UK \\
\email{\{o.cory,o.mercanoglusincan,r.bowden,m.j.vowels\}@surrey.ac.uk}\and
The Sense Innovation and Research Center, Lausanne and Sion, CH \and
Institute of Psychology, University of Lausanne (UNIL), Lausanne, CH \and
Department of Computational Linguistics, University of Zurich, Zurich, CH \\
\email{\{alessia.battisti,sarah.ebling\}@uzh.ch} \and
University of Teacher Education in Special Needs (HfH), Zurich, CH \\
\email{\{franz.holzknecht,katja.tissi\}@hfh.ch}, \email{sandysidler@gmail.com}}

\maketitle

\begin{abstract}
  Sign Language Assessment (SLA) tools are useful to aid in language learning and are underdeveloped. Previous work has focused on isolated signs or comparison against a single reference video to assess Sign Languages (SL). This paper introduces a novel SLA tool designed to evaluate the comprehensibility of SL by modelling the natural distribution of human motion. We train our pipeline on data from native signers and evaluate it using SL learners. We compare our results to ratings from a human raters study and find strong correlation between human ratings and our tool. We visually demonstrate our tools ability to detect anomalous results spatio-temporally, providing actionable feedback to aid in SL learning and assessment.

\end{abstract}

\section{Introduction}
\label{sec:intro}

Sign Languages (SL) are nuanced and complex visual-gestural languages that are the primary form of communication for millions of deaf \footnote{We follow the recent convention of abandoning a distinction between \emph{Deaf} and \emph{deaf} and use the latter term also to refer to (deaf) members of the sign language community \cite{kusters-et-al-2017, napier-leeson-2016}.} people worldwide. With the advancements in deep learning and computer vision, there has been a growing interest in modelling SL. The majority of methods focus on classification, namely for the recognition and translation of sign \cite{KollerO20160626-20160701DHHt, 8578910}, rather than improving or assessing SL proficiency. The standardisation of Sign Language Assessment (SLA) is a challenging research topic due to the many nuances that affect its legibility \cite{Haug2022Handbook}.

The study of Sign Language Linguistics is still in its infancy, especially when compared to spoken languages. SL have no standardised written form, they are conveyed via a combination of manual and non-manual features \cite{https://doi.org/10.1111/j.1467-968X.2010.01242.x}. While the manual features include the location, orientation, and movement of the arms and hands; non-manual features refer to facial expressions, body posture, head movement, and eye gaze. Signing involves simultaneous combinations of these features, each influencing the meaning of a sign, adding multiple layers of linguistic complexity. In continuous sequences, co-articulation is also common factor \cite{sandler2006sign}. This includes temporal overlap between signs in a sequence leading to blending, spatial influence where the location of one sign may impact the starting location of the following signs, and handshape modifications based on context. Given the rich and complex nature of SL, skilled teachers are needed to assess and quantify signing proficiency.

In this paper we focus on SL assessment, proposing a tool to aid human teachers to evaluate continuous SL and to improve efficiency in evaluation and feedback. Teaching systems for SL that incorporate feedback mechanisms have been proposed using classification to determine correct from incorrect repetitions or to regress scores directly \cite{TornaySandrine20201025, wen2024learning}. However, most approaches are limited to the assessment of isolated signs \cite{Tornay2023WebSMILEDemo}.

Our work provides an SL assessment tool for continuous sequences that learns the natural distribution present in human motion. We develop a \emph{Skeleton Variational Autoencoder (SkeletonVAE)} to embed signed sequences from multiple native signers in a compact, lower dimensional subspace. We then apply a \emph{Reference Selection} technique over these embeddings to determine the most representative sequence from the collection of sequences. We finally model the \emph{Motion Envelope} by aligning all the sequences to the reference and learning the distribution over the embedded data using a Gaussian Process (GP). 

We test our model using data from SL learners and evaluate its performance against ratings collected from a human raters study. We demonstrate that our model can quantitatively evaluate the production of sequences achieving similar results to a manual rater. Furthermore, we show that our system can determine where and by what distance a learner falls outside of the natural acceptable variation in human motion for signed sequences.

\section{Related Work}
\subsubsection{Sign Language Recognition, Translation and Production.}
Computational approaches to SL modelling have been the focus of researchers for over 30 years \cite{TAMURA1988343}. Preliminary research focused on isolated Sign Language Recognition (SLR) using statistical methods \cite{625742}, aiming to classify isolated signs. The advent of deep learning techniques has enabled the development of continuous SLR methods operating over continuous sequences, implemented using CNNs \cite{KollerO20160626-20160701DHHt, KollerO20160919-20160922DSHC}, RNNs \cite{inproceedings, Cui2017RecurrentCN}  and more recently Transformers \cite{Camgoz_2020_CVPR}. Some researchers operate in the pixel space directly whereas others choose to use skeleton pose, optical flow, or a combination of modalities \cite{jiang2021skeleton,selvaraj2021openhands}. 

More recently the field has moved towards Sign Language Translation (SLT) \cite{8578910}, aiming to translate continuous sign to written spoken language sentences rather than just recognising the consistent signs. SLT is a more challenging task than SLR due to the grammatical and ordering differences between SL and spoken language. Transformer-based approaches have achieved state-of-the-art performance, learning the recognition and translation tasks jointly \cite{Camgoz_2020_CVPR}. 

Most SLT approaches require intermediate representations and while traditional approaches often rely on linguistic representations such as gloss \cite{8578910, Camgoz_2020_CVPR} or HamNoSys \cite{KollerO20160523-20160528AAoH, walsh2022changing}, such annotation is expensive to create. To overcome this bottleneck, recent research has shifted towards gloss-free translation \cite{wong2024sign2gpt, Gong_2024_CVPR, yin2023gloss,zhou2023gloss, walsh2024datadriven}.

On the other hand, Sign Language Production (SLP) aims to produce SL videos from written spoken language sentences. Current approaches to SLP use Transformer-based architectures, extending SLT to include the production of digital avatars \cite{kim-etal-2022-sign}, photo-realistic outputs using Generative Adversarial Networks \cite{saunders2020progressive, stoll2020text2sign} or diffusion models\cite{fang2023signdiff}.

\subsubsection{Language Learning and Assessment.}
Automated tools for language learning have been widely developed for written and spoken languages. Mainstream tools such as Duolingo \cite{duolingo} have proven their effectiveness in increasing learning efficiency through gamification. There are only a few studies that utilize gamification for SL which aim to teach isolated signs \cite{starner2023popsign, 10.1145/3411763.3451523}.

Sign Language Assessment (SLA) systems that provide more detailed feedback have also been introduced \cite{10.1145/3536221.3556623, Tornay2023WebSMILEDemo, TornaySandrine20201025}. Tornay et al.\cite{Tornay2023WebSMILEDemo} provide a scoring mechanism alongside a visualisation showing the performed sign against a reference skeleton, providing actionable feedback. However, this is limited to isolated signs. Wen et al. \cite{wen2024learning} introduced an approach for SL assessment over continuous sequences using a two-stage method that integrated domain knowledge from action similarity techniques. However, the method relies on a single reference video to evaluate against and therefore does not account for the natural variation in human motion during assessment.

The complexity of SL makes its annotation and assessment challenging. Annotating SL data is extremely time consuming, with one minute of sign taking between 10 and 30 minutes to annotate \cite{kindiroglu2024transfer}. The natural variability in signing between individuals (often referred to as signer `style' \cite{krebs-etal-2024-motion-capture}) further complicates data annotation and quantification of SL proficiency. Human assessment remains the most reliable method for scoring SL, as teachers can accurately determine correct sign production despite natural variation between signers \cite{4813347}. Holzknecht et al. \cite{franz_assessment_june_24} compare the results of an automated SLA system with ratings from a human rater study for isolated signs. We compare our approach to human raters in the context of continuous signed sequences.

\subsubsection{Action Quality Assessment.}
The assessment of SL can be seen as a subdomain within the broader field of Action Quality Assessment (AQA), which aims to evaluate the quality and performance of human actions in various contexts. Previous research in AQA has typically focused on macro-level actions \cite{jain2020action, xu2022finediving, morais2019learning} rather than fine details like individual hand and finger movements. The majority of these methods compare actions against a single reference and directly regress a score\cite{xu2022finediving, jain2020action}. Some use additional hardware to track human behaviours \cite{longfei2020modeling}, some work directly from RGB videos \cite{xu2022finediving}, and others use pose representations \cite{morais2019learning, wen2024learning}.

Morais et al. \cite{morais2019learning} employ a pose representation for anomaly detection. However, this method requires high correspondence of movements to be effective, with performance degrading due to the natural variations in posture between individuals. Xu et al. \cite{xu2022finediving} rely on meticulously annotated training data to achieve high accuracy results , making it a highly supervised approach.

Two-stage approaches have emerged offering a more flexible and interpretable framework for assessing action quality by decoupling feature extraction from the evaluation process, allowing for more adaptable and insightful AQA systems \cite{taichi, jainunsuper, wen2024learning}. In the domain of SL, Wen et al. \cite{wen2024learning} proposes a two-stage pipeline where features are first recovered from video, embedded and aligned to a single reference using Dynamic Time Warping \cite{salvador2007toward} before assessing sign quality. 

Distinct from previous approaches, our unsupervised method accounts for the natural variation in human motion when assessing action quality by learning the distribution in motion over multiple expert productions.

\section{Method}
\begin{figure}[ht]
  \centering
  \includegraphics[width=1\textwidth]{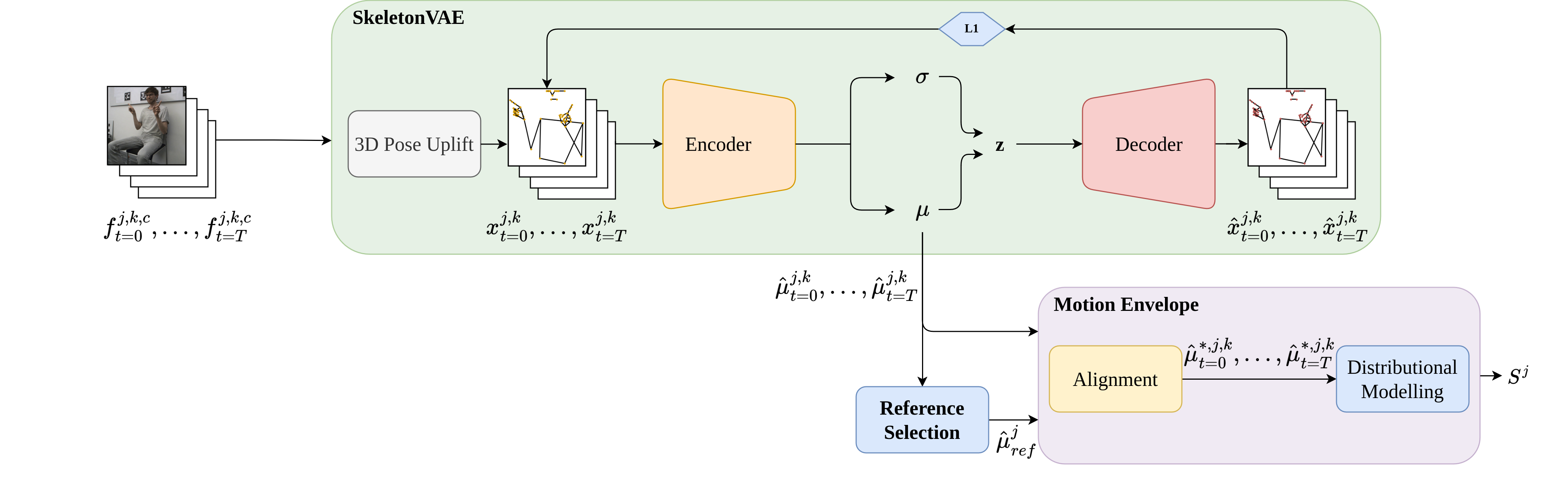}
  % \includesvg[width=\linewidth]{arch_diagram.svg}
  \caption{Diagram showing training pipeline for modelling the $j^{th}$ sentence over K signers. The process takes J example sentences captured with C independent cameras and uses 3D pose uplift to create a set of $\mathbf{x}$ poses which are fed into the VAE, encoding the poses into $\boldsymbol{\hat{\mu}}$ latent means. Reference Selection finds the central signal $\hat{\boldsymbol{\mu}_{ref}}$ and learns a distribution over K signers.}
  \label{fig:arch_diag}
\end{figure}

We present a novel approach for learning the natural distribution of continuous Sign Language sequences. We first build a \emph{SkeletonVAE} by uplifting multi-view video data to a 3D skeleton pose and learning a low-dimensional latent representation of pose, capturing the essential characteristics of human movement. We take our video dataset of sentences with multi-participant productions and encode them to create a secondary dataset of latent time varying embeddings. Second, we develop a  \emph{Reference Selection} technique which identifies a reference production of each sentence based on similarity calculation between all participants. Finally, we build a \emph{Motion Envelope} by aligning each participant's sequence to its corresponding sentence reference and model the distribution of per-dimension embedding trajectories across multiple signers. The pipeline for this method is shown in Fig. \ref{fig:arch_diag}.

\subsection{SkeletonVAE}

Consider $N$ sequences of SL video frames $\mathbf{f}^{j,k,c}_{t}$, where $j=\{1, 2..., J\}$ sign language sentences being executed by $k = \{1, 2,...,K\}$ individual signers, and where $t$ is an individual timepoint ranging $t = \{1, 2,...,T_{j,k}\}$, such that the total number of timepoints depends on the signer and the sentence being performed. $c$ indexes $C$ synchronised cameras that capture all the data together.

We start by extracting Mediapipe \cite{lugaresi2019mediapipe} 2D poses from a single view for $C$ cameras over the entire dataset. After this, we implement 3D pose uplift \cite{ivashechkin2023improving} to regress accurate 3D skeleton data and convert to canonical form by choosing fixed bone lengths and applying this scaling via the joint angles. We now have $N$ sequences of $d$-dimensional skeleton joint-position data for sign language poses $\mathbf{x}^{j,k}_{t}$.

We assume that, within the context of human motion and SL, each pose lies on some manifold with fewer dimensions than $d$ which we can approximate via a stochastic mapping $p_{\theta}(\mathbf{z} | \mathbf{x}_t^{j,k}) : \mathbf{x} \rightarrow \mathbf{z}$ where $\mathbf{z} \in \mathbb{R}^{\Omega}$ is a latent representation or embedding. Our goal is to model the time variation of $\mathbf{x}$ in terms of its compact representation $\mathbf{z}$.

We begin by taking the skeleton poses $\mathbf{x}$ and embedding them using a Variational AutoEncoder, which is trained via a process known as variational inference \cite{Blei_2017, higgins2017betavae, kingma2022autoencoding}. Variational inference is concerned with maximising the Evidence Lower BOund (ELBO), which forms a lower bound on the negative log-likelihood of the data under the model:

\begin{equation}
\begin{split}
   N^{-1} \sum_{i=1}^{N} \log p_{\theta} (\mathbf{x_i}) \leq \\
   N^{-1} \sum_i^N \left( -\mathbb{E}_{q_{\phi}(\mathbf{z}|\mathbf{x}_i)} \left[ \log p_{\theta}(\mathbf{x}_i|\mathbf{z})\right] + \beta \mathbb{D}_{KL}\left[q_{\phi}(\mathbf{z}|\mathbf{x}_i)||p(\mathbf{z})\right] \right) \;.
    \end{split}
    \label{eq:elbo}
\end{equation}

Here, $q_{\phi}(\mathbf{z}|\mathbf{x})$ is known as the approximating posterior, which ideally matches the true posterior $p(\mathbf{z}|\mathbf{x})$ which we do not have access to. We therefore assume a parameterisation for this approximating posterior, and define a prior distribution $p(\mathbf{z})$. The Kullback-Liebler divergence $\mathbb{D}_{KL}$ is then used to create pressure such that the approximating posterior distribution $q$ resembles this prior, and this pressure is weighted with a scalar $\beta$ \cite{higgins2017betavae}. Using a $\beta$ value other than one means the ELBO cannot technically be fulfilled, but is a hyperparameter determined by experimental results. For our work the choice of prior is an isotropic Gaussian with mean $\boldsymbol{\mu}=0$ and variance $\boldsymbol{\sigma}^2=1$. The parameters $\phi$ and $\theta$ represent the neural network parameters for the encoder and decoder respectively, and it is the encoder and decoder which parameterise the approximating posterior and conditional likelihood models. As such, each datapoint is encoded as a mean $\hat{\boldsymbol{\mu}}$ and a variance $\hat{\boldsymbol{\sigma}^2}$ which, via the reparameterisation trick \cite{kingma2015variational}, enable sampling $\mathbf{z}|\mathbf{x}_i \sim \mathcal{N}(\hat{\boldsymbol{\mu}}_i, \hat{\boldsymbol{\sigma}^2_i})$ which are decoded to reconstructions $\hat{\mathbf{x}}_i$. Finally, note that in Eq.~\ref{eq:elbo}, $N$ is the total number of datapoints available, across all $J$ sentences and $K$ signers (i.e. $N=K \times J \times T$ for fixed $T$). This part of the modeling process therefore treats the data as independent and identically distributed (the sequential aspect of the data, as well as the fact we have different sentences being performed, will be modeled using Gaussian Processes).

\begin{equation}
    \mathbf{\mathcal{L}_{VAE}} = \alpha\mathbf{{L1}_{hands}} + (1-\alpha)\mathbf{{L1}_{body}} + \mathbf{\beta\mathbb{D}_{KL}}
    \label{eq:vae_loss}
\end{equation}

Since hands are high-frequency, low-amplitude signals due to their rapid and detailed movements compared to the larger, slower movements of the body, they can be lost in the noise during VAE training. To address this, we use L1 loss as the reconstruction loss and split the weighting of the loss between the hands and body. By setting a high $\alpha$ value, the network can better focus on hand reconstruction. Our overall loss function, \cref{eq:vae_loss}, is the sum of this reconstruction loss with the $\beta$-scaled KLD. Once we have trained the VAE on all skeleton poses for the complete dataset, we arrive at a secondary dataset of encodings $\hat{\boldsymbol{\mu}}_{t}^{j,k}$ where $\hat{\boldsymbol{\mu}}$ represents the conditional mean encoding of the corresponding skeleton datapoint $\mathbf{x}$.

\subsection{Reference Selection}
For $J$ sentences, we calculate a cosine similarity matrix comparing the encoded means $\hat{\boldsymbol{\mu}}$ over $K$ signers. We then average the matrix entries for each $k$, returning the average similarity scores of $k$ with reference to all other $k$'s that produced $j$. We choose the highest average similarity signal as our reference signal $\hat{\boldsymbol{\mu}}_{ref}^j$, for each sentence. This signal is the central signal and is used as the reference for the Dynamic Time Warping (DTW) \cite{salvador2007toward} algorithm.

\subsection{Motion Envelope}
At this stage we use DTW to align the sequences such that $T_{j,k} = T^*_j \forall k$, where $T^*_j$ is the length of the reference sequence $\hat{\boldsymbol{\mu}}_{ref}^j$. Each of the aligned sequences are denoted $\hat{\boldsymbol{\mu}}^*$. 

Finally we train a Gaussian Process (GP) \cite{10.7551/mitpress/3206.001.0001} for each of the $J$ sentences, for  $\hat{\boldsymbol{\mu}}^*$, across the $K$ individual signers. In other words, we take time-aligned sequences for a particular sentence and train the GP using the multiple productions of that sentence by the $K$ signers. The trained model for a specific sequence $j$ is denoted as $\boldsymbol{S}_j$:

\begin{equation}
\boldsymbol{\hat{\mu}}_t^{*, j} \sim \boldsymbol{S}_j: = GP\left(mean^{j}(t), cov^{j}(t, t') \right),
\end{equation}

where $mean$ is a mean function and $cov$ is a covariance function determining the covariance between any pair of timepoints $t$ and $t'$. The GP therefore provides us with an approximation of the distribution of embedding trajectories for a particular sentence, across multiple signers.

To train the GP models, we utilise the negative of the marginal log likelihood (MLL) as our loss function. The negative MLL for the aligned latents $\boldsymbol{\hat{\mu}}^{*, j}$ given the inputs $T_j$ is defined as:

\begin{equation}
\begin{split}
\boldsymbol{\mathcal{L}_{GP}} = -\log p(\boldsymbol{\hat{\mu}}^{*, j} \lvert T^*_j)\\
= -\log \mathcal{N}(\boldsymbol{\hat{\mu}}^{*, j} \lvert mean^j, {cov}^j) \\ 
= -\frac{1}{2} \left(\boldsymbol{\hat{\mu}}^{*, j}-mean^j\right)^T \left({cov^j}\right)^{-1} \left(\boldsymbol{\hat{\mu}}^{*, j}-mean^j\right) 
-\frac{1}{2} \log \begin{vmatrix}cov^j\end{vmatrix} 
-\frac{N}{2} \log(2\pi)
\end{split}
\end{equation}

where $N=K_jT^*_j$, and $K_j$ is the number of signers that produced sentence $j$. By taking the negative of the MLL, we maximise the log likelihood of the observed data under the GP model, thereby fitting the model to the data in a way that best explains the observed latents $\boldsymbol{\hat{\mu}}^{*, j}$.

At inference time we take the embeddings for a test sequence for a specific $j$, $\hat{\boldsymbol{\mu}}_{test, j}$ and align it to the corresponding $\hat{\boldsymbol{\mu}}_{ref}$ such that it becomes $\hat{\boldsymbol{\mu}^*}_{test, j}$. We compare this sequence to the multivariate Gaussian posterior of the learnt model $\boldsymbol{S}_{j}$, returning principled uncertainty estimates for each $t$ in the sequence.

\section{Experiments}
We evaluate our method using real-world SL data from native signers and language learners. We first outline our SL Sentence Repetition Test dataset and discuss the human rating scheme. We provide implementation details and compare our approach to the manual ratings by demonstrating quantitative and qualitative results. 
\subsection{Dataset}

\begin{figure}[ht]
  \centering
  \begin{subfigure}{0.32\linewidth}
    \includegraphics[width=\columnwidth]{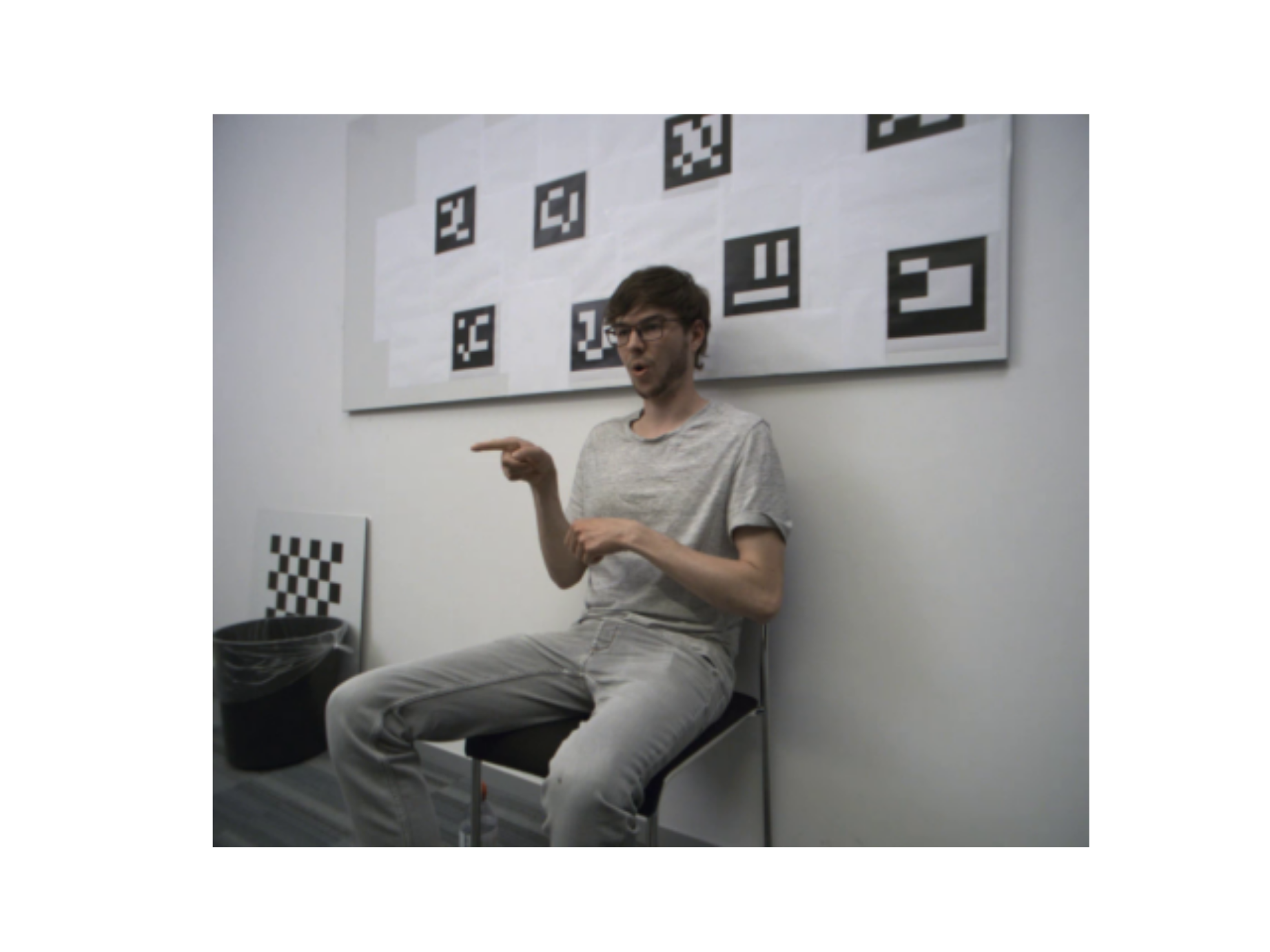}
    \caption{RGB}
        \label{fig:rgb_img_example}
  \end{subfigure}
  \hfill
  \begin{subfigure}{0.32\linewidth}
    \includegraphics[width=\columnwidth]{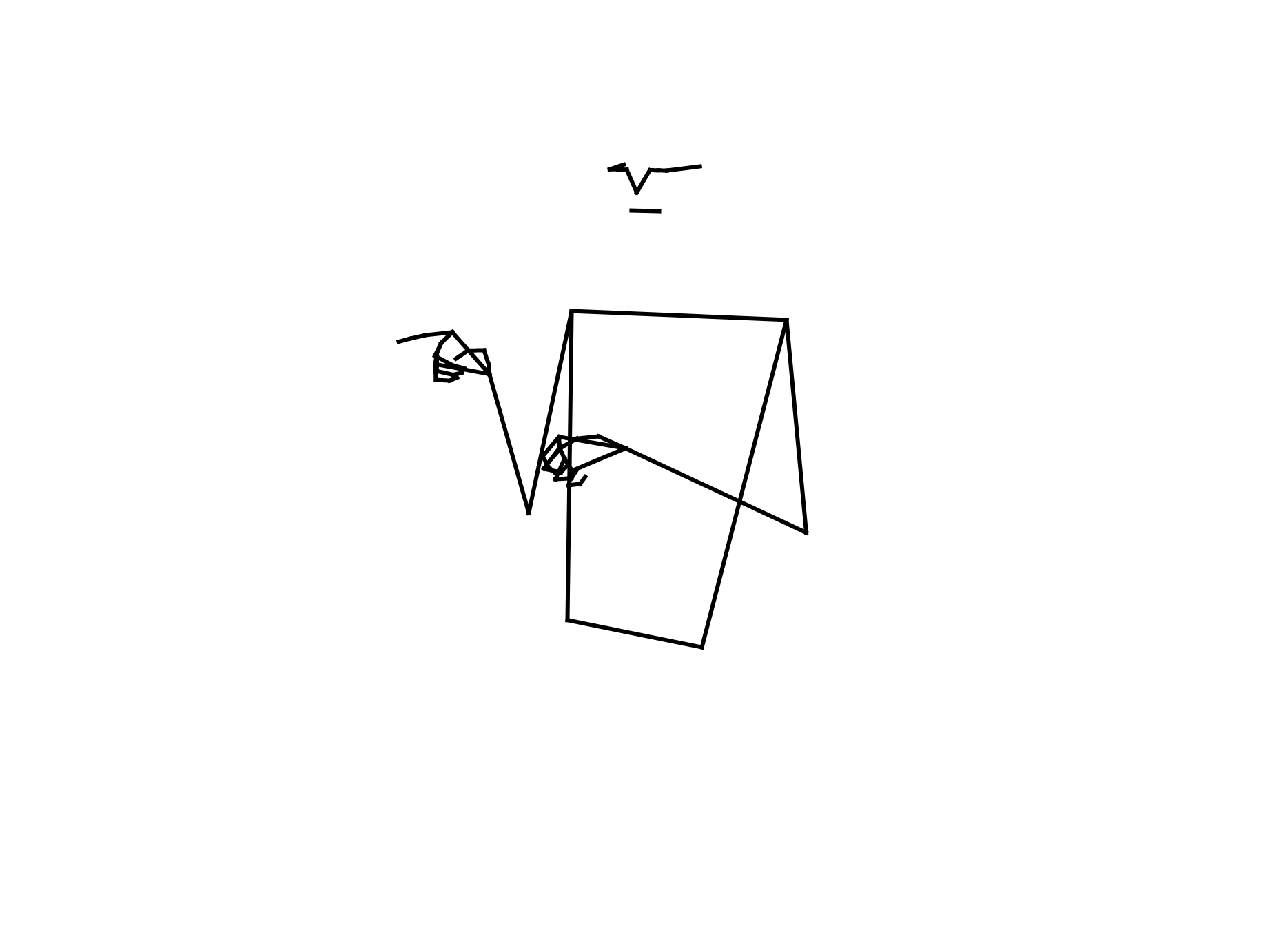}
    \caption{Extracted 3D Skeleton}
        \label{fig:extract_skel_example}
  \end{subfigure}
  \hfill
  \begin{subfigure}{0.32\linewidth}
    \includegraphics[width=\columnwidth]{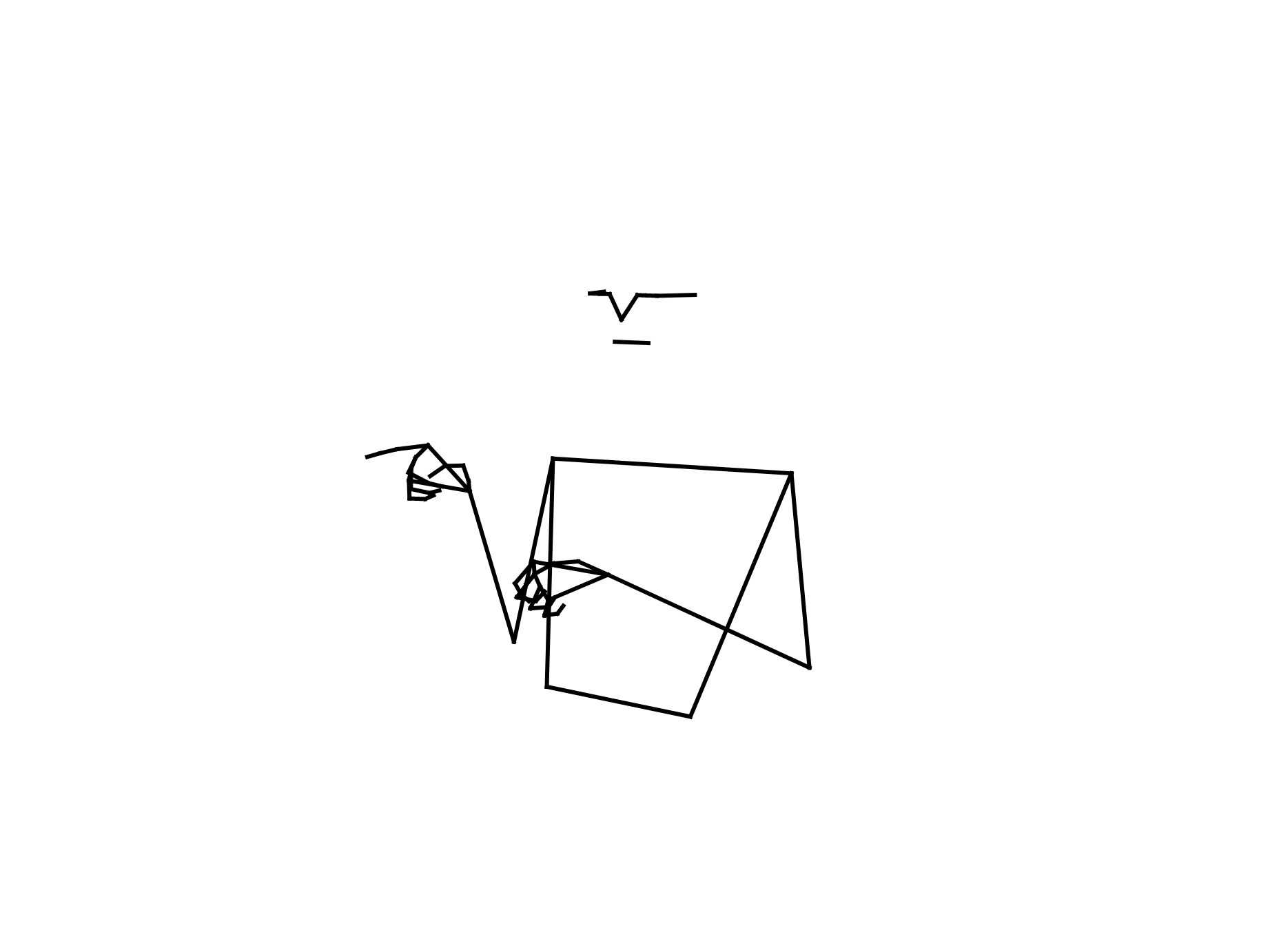}
    \caption{Canonical Skeleton}
        \label{fig:canon_skel}
  \end{subfigure}
  \caption{Example frame from the dataset showing \ref{fig:rgb_img_example} the RGB frame of a participant from one of the camera views, \ref{fig:extract_skel_example} the uplifted 3D skeleton, and \ref{fig:canon_skel} the bone length adjusted canonical skeleton.}
  \label{fig:learner_mean_native_positive}
\end{figure}

A recent study suggests that Sentence Repetition Tests (SRTs), which are widely used as a means of assessment for spoken language, can be applied to SL assessment \cite{doi:10.1177/0265532219898382}.
SRTs ensure a comprehensive evaluation of signing ability; by requiring both comprehension and production, they provide a robust measure of language proficiency in the context of SL.
During the testing process, each participant sees a prerecorded signed sequence video twice and is then asked to repeat it, i.e., the test taker has to comprehend, process, and produce language \cite{doi:10.1177/0265532215594643}. SRTs often work with a binary concept of correctness \cite{doi:10.1177/0265532215594643}. In this work a partial credit scale is used (as in \cite{https://doi.org/10.1111/1473-4192.00024}) in order to provide more informative feedback to the participant.

\begin{table}[ht]
\centering
\caption{Selection of examples from the Sentence Repetition Test}
\label{tab:SRT_sentences}
\begin{tabular}{lp{30em}r} % Adjusted column width
\toprule
{Sentence ID} & {High German Written Sentence (English Translation)} \\
\midrule
A       & Das Essen gestern Abend im Restaurant war schlecht.\newline (Last night the food in the restaurant was bad.) \\
E       & Ich mag diesen Salat gar nicht. \newline (I don't like this salad at all.) \\
L       & Er/Sie ist nicht da, weil er/sie krank ist. \newline (He/she is not there because he/she is sick.) \\
\bottomrule
\end{tabular}
\end{table}

We create our Swiss-German Sign Language (Deutschschweizerische Gebärdensprache, DSGS) SRT dataset by recording a repetition test across 12 sentences of varying difficulty, determined by the number of signs, as well as morphological and syntactic complexity. The test is taken by a combination of 10 native signers and 14 language learners. Some examples of the sentences are shown in \cref{tab:SRT_sentences}. We use the data from the native signers as our gold standard for training our model and the learners' data for evaluation.

We extract 3D canonical skeleton pose data (as shown in \cref{fig:canon_skel}) for the dataset, with each pose represented by 61 nodes in 3D Cartesian space. We sort the native signer data to include only sentences which are produced in the sentence order matching the initial reference and use this data for training the GPs model. We evaluate the model using all sentences produced by the language learners.

\subsection{Manual Ratings}
The data is analysed by eight native raters of DSGS using rating criteria designed to provide a comprehensive assessment of signing accuracy and fluency \cite{franz_rating_smile_ii_pres}. Raters are trained on a standardised rubric and evaluate videos of the sentences across six criteria: manual components, mouth components, eyebrow movements, head movements, eye gaze, and sentence structure.

Each criterion is assessed on a three-point scale, allowing for more nuanced feedback compared to a binary system. To ensure reliability, 14 videos are designated as anchor videos and are rated by all eight raters. The remaining 97 videos are assessed by two raters each in an overlapping design, with measures taken to balance video allocation and minimise potential bias. Analysis of the ratings provides inter-rater reliability, allowing us to determine the most reliable criterion for assessment.

For our experiments we choose to evaluate against the criterion for manual features on the sign produced by the language learners. Each rater provides a score for each manual component of the sign in the sentence. We take the mean of the ratings across the components in the sentence for each rater; and then take the mean across the raters that rated the sentence-learner pair. We repeat this for every sentence and learner. This provides a single score, 1 to 3, for each learner, for each sentence, that can be used for comparison with the output of our system.

\subsection{Implementation Details}
The encoder of our VAE consists of an input layer of size 183, followed by two hidden layers with sizes 100 and 50 perceptrons respectively. We implement fully connected layers and ReLU activation functions at the output of each layer except the final output layer, where we use a TanH function with its output scaled by a value of 6 to map the output of the network to the coordinate space of our pose data. The output of the encoder is split into two separate fully connected layers, each of size 10, representing the mean and log-variance of the latent space distribution. The mean and log-variance values are combined using the reparameterisation trick to calculate a 10-dimensional z-value vector. The decoder mirrors the structure of the encoder. It takes the 10-dimensional latent vector and passes it through two hidden layers, of size 50 and 100 respectively. The final output layer of the decoder reconstructs the original input dimension with size 183.

We initialize the weights of the fully connected layers using Kaiming normal initialization \cite{he2015delving}, with the biases initialized to 0.01. During training, we scale the added noise by a value of 0.001. For our loss function, \cref{eq:vae_loss}, we choose an $\alpha$ of 0.9 and a $\beta$ value of 0.0001 based on empirical experimentation. We train with a batch size of 32 for 100,000 epochs, with a learning rate of 0.001. We use Adam as our optimizer \cite{kingma2017adam}, and train over all canonical skeletons in the dataset.

For the DTW we choose a radius of size 20. For our GP Regression model we implement the `ExactGP' model from GPyTorch \cite{gardner2018gpytorch}. We choose Gaussian likelihood as our likelihood function, use a Radial Basis Function as our kernel type, and initialise the mean function as a constant set to zero. We implement a gamma prior over the length scale with concentration and rate values both set to 0.1. We train with a learning rate of 0.1 until the loss reaches a threshold of 0.001.

\subsection{Quantitative Results}
In this section we evaluate the performance of our system using two distinct methods. 
The first method, which we refer to as the Probability Density method (PD method) is as follows. For each $t$ in the test sequence, we calculate the probability density of the learner with respect to the learnt distribution at that point in the sequence for each latent dimension, resulting in a multidimensional mean. We then take the average of this mean, resulting in a single score for signing proficiency which we refer to as the Probability Density Measure (PD Measure). We expect a learner assigned a high manual rating to receive a high PD Measure and vice versa.

The second method quantifies the number of instances where the learner deviates from the distribution defined by the Motion Envelope, we refer to this as the Out of Distribution Count. Specifically, this method counts the occurrences where the learner falls outside the high confidence region, summing across all dimensions. The high confidence region is defined as the region that covers where we expect the true function values to lie with 95 percent probability \cite{10.7551/mitpress/3206.001.0001}. This method is particularly effective at assessing anomaly detection. We expect a learner assigned a high manual rating to receive a low Out of Distribution Count and vice versa.

We standardise the resulting scores from our model and the manual ratings data using z-scoring standardisation. We apply the standardisation across each rater individually for all their ratings which increases the comparability of ratings from different raters. We then apply the standardisation on a per sentence level for the output scores of our system and the manual ratings. This results in standardised beta coefficients (which range from -1 to +1) when performing the regression analysis.

\subsubsection{Linear Regression Analysis.}
We first evaluate our system by performing linear regression between the output scores of our model and the manual ratings data, measuring the standardised beta coefficient.

\begin{table}[ht]
\centering
\caption{Linear Regression Results. \emph{Bolded} results for $\beta$, the \emph{Standardised Beta Coefficient}, indicate the stronger correlation for each sentence out of the two methods. $\beta$ represents the degree of correlation between the manual ratings and the outputs of the system.}
\label{tab:linear_regression_results}
\begin{tabular}{l@{\hspace{1.5em}}r@{\hspace{1.5em}}r}
\toprule
{Sentence} & {Prob. Density Measure $\uparrow$} & {Out of Dist. Count $\downarrow$} \\
\midrule
A & \bfseries0.60  & -0.15  \\
B & \bfseries0.19  & 0.03  \\
C & \bfseries0.31  & -0.28  \\
D & 0.37  & \bfseries-0.40  \\
E & \bfseries0.18  & -0.09  \\
F & 0.35  & \bfseries-0.70  \\
G & \bfseries0.37  & -0.35  \\
H & 0.27  & \bfseries-0.45  \\
I & \bfseries0.24  & -0.09  \\
J & 0.00  & \bfseries-0.36  \\
K & \bfseries0.57  & -0.17  \\
L & \bfseries0.46  & -0.45  \\
\bottomrule
\end{tabular}
\end{table}

\begin{table}[!ht]
\centering
\caption{Spearman Rank Correlation Coefficient Results. \emph{Bolded} results indicate the stronger correlation for each sentence out of the two methods.}
\label{tab:spearman_results}
\begin{tabular}{l@{\hspace{1.5em}}r@{\hspace{1.5em}}r}
\toprule
{Sentence} & {Prob. Density Measure $\uparrow$} & {Out of Dist. Count $\downarrow$} \\
\midrule
A       & \bfseries0.69 & -0.19 \\
B       & \bfseries0.27 & -0.20 \\
C       & 0.31 & \bfseries-0.49 \\
D        & 0.35 & \bfseries-0.42 \\
E        & \bfseries0.30 & -0.15 \\
F        & 0.43 & \bfseries-0.60 \\
G        & 0.38 & \bfseries-0.50 \\
H        & 0.22 & \bfseries-0.51 \\
I        & \bfseries0.20 & 0.14 \\
J        & -0.03 & \bfseries-0.43 \\
K        & \bfseries0.56 & -0.26 \\
L        & \bfseries0.44 & -0.40 \\
\bottomrule
\end{tabular}
\end{table}

The results for the two methods can be seen in \cref{tab:linear_regression_results}. A notable result here is the difference in scores when assessing using the PD Measure or Out of Distribution Events. For sentences A, B, C, E, G, I, K, L the first method achieves the best results, where as for sentences D, F, H, J the second method performs better. Both methods can be deemed useful. The PD method provides a more complete score over the entire sequence as all points in time are used in its calculation. However, it may be skewed negatively by acceptable deviations in sentence productions that are within distribution but far from the mean, as these will score relatively low compared to those with smaller distances to the mean of the distribution.

The Out of Distribution Count method only incorporates events into the score when the threshold is exceeded, providing a good method for anomaly detection, countering the downside of the PD method mentioned above. 

For some of the sentences, the results for the PD measure and Out of Distribution Count are both low. One reason for this may be due to a non-linear relationship between the manual ratings and the output of the system. To investigate this we present results using the Spearman Rank Correlation Coefficient.

\subsubsection{Spearman Rank Correlation Coefficient (SRCC).}
The SRCC is a measure of the strength and direction of the association between two variables that are assumed to be monotonic but not necessarily linear, based on the ranked values of the data.

In \cref{tab:spearman_results} we show that this metric offers complementary validity to that in \cref{tab:linear_regression_results} suggesting that the results are robust and not a spurious outcome of metric choice. Furthermore, the strong SRCC scores shows that the monotonic relationship between the manual ratings and the system scores may be non-linear.

\subsection{Plot Analysis} 
\begin{figure}[hb]
  \centering
  \includegraphics[width=0.6\textwidth]{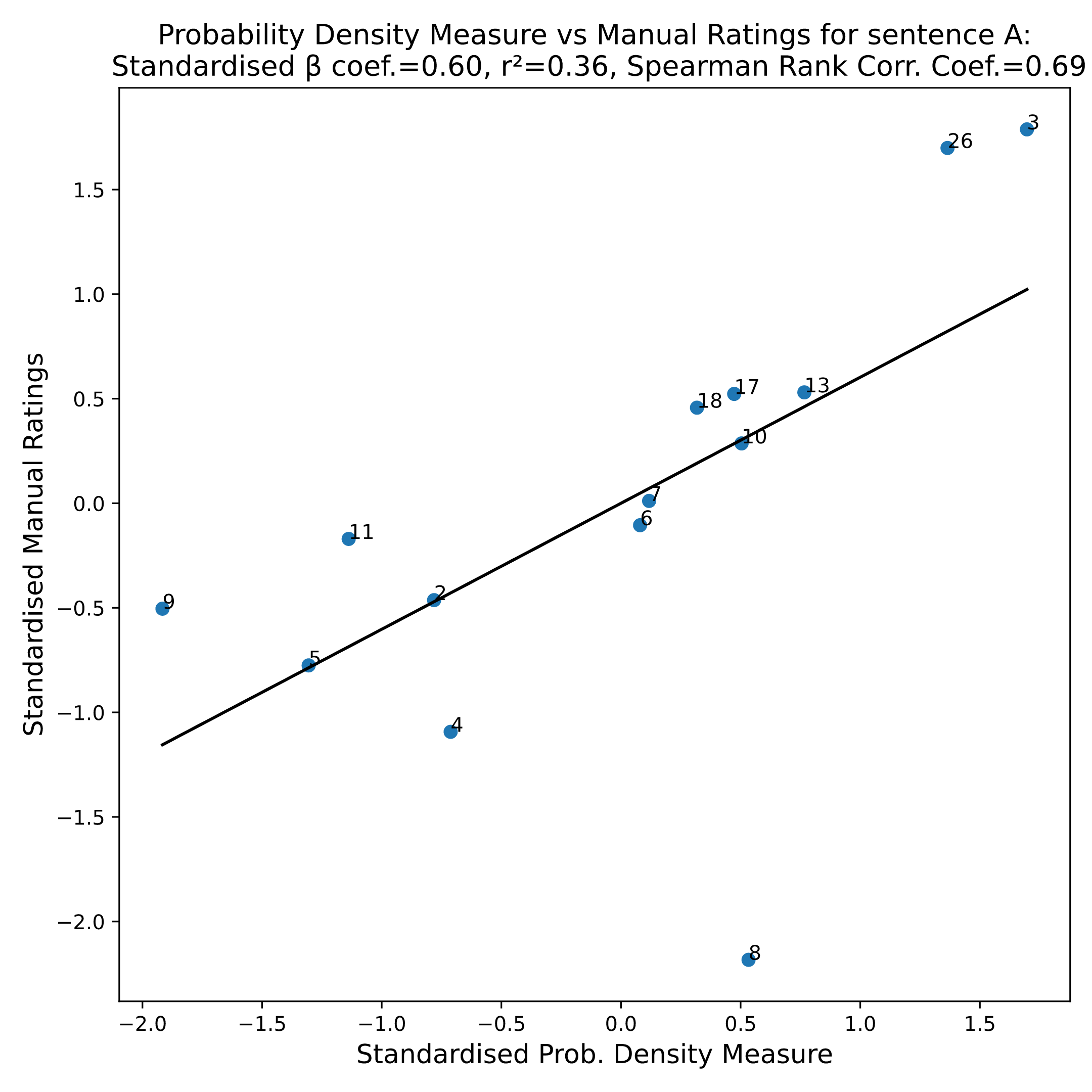}
  \caption{Figure showing \emph{standardised PD Measures} against the \emph{standardised manual ratings} for \emph{sentence A}. The \emph{blue points} represent the language learners that produced the sentence, labelled with their predefined signer ID. The \emph{black line} represents the line of best fit from the linear regression.}
  \label{fig:sent_12_single_score}
\end{figure}

\cref{fig:sent_12_single_score} showcases the model's strong agreement with the manual rating data for an example sentence. The model assigns low and high scores to the correct learners with respect to the manual ratings data, demonstrating its effectiveness in SL assessment. The correlation is strongly positive and almost linear.

Learner 8 is a significant outlier in this plot, with our system assigning a mid-level score but being manually rated low. When looking at the Many Facets Rasch Measurement \cite{myford2003detecting} for severity among raters, it becomes apparent that the sample is an outlier due to it being rated by the two most severe raters. In this case, the manual rating may be skewed negatively by their severity.

\subsection{Qualitative Results}
We now examine our system qualitatively, by using examples of a high and low scoring SL learner with respect to the learnt Motion Envelope and visualise their results.

\begin{figure}[tb]
    % \includesvg[width=1.0\textwidth]{sentence_12_dim_1_start_100_end_250.svg} SVG BROKEN
    \includegraphics[width=\textwidth]{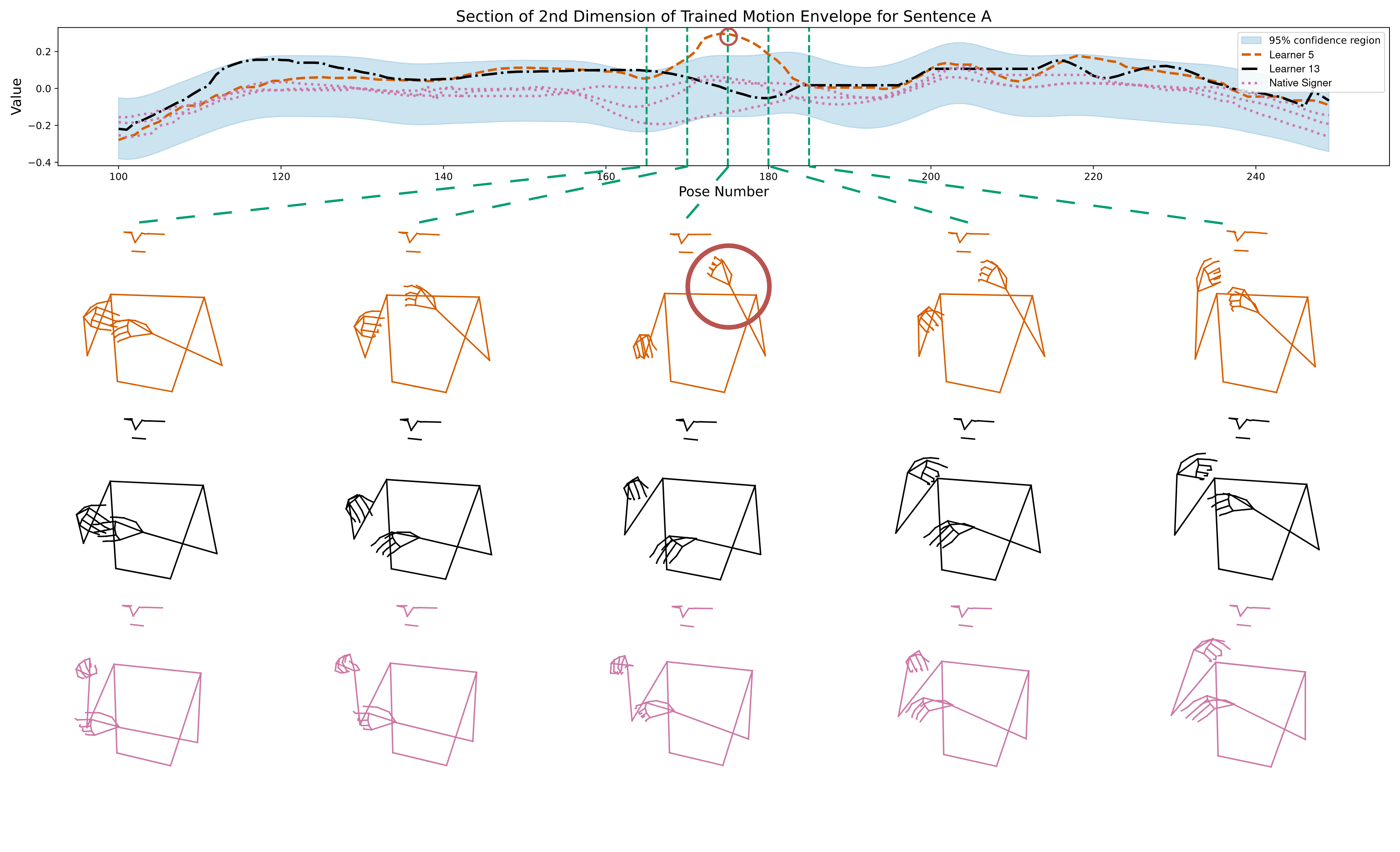}
    \caption{Top plot shows a section of from the latent dimension of the Motion Envelope \emph{Confidence Region} with encoded SkeletonVAE signals overlayed for \emph{Sentence A}. Below, decoded pose data for the latents is visualised for \emph{Learner 5} (top), \emph{Learner 13} (middle) and one \emph{Native Signer} (bottom) for Pose Numbers 165-185 in steps of 5. The \emph{red circle} indicates \emph{Learner 5}'s peak deviation from the distribution.}
    \label{fig:dist_figure}
\end{figure}

 As shown in \cref{fig:sent_12_single_score}, Learner 5 and Learner 13 both lie close to the line of best fit of the linear regression. Learner 5 receives a low overall single score from our system and is similarly rated by the manual raters whereas for Learner 13 the opposite is true, receiving high scores. As such these two language learners make a good example for further evaluation. 

The plot on \cref{fig:dist_figure} shows time varying latent signals from one of the SkeletonVAE dimensions for Sentence A ranging from Pose Number 150 to 250 for learners and native signers against the learnt Motion Envelope confidence region. Learner 5 is shown leaving the learnt confidence region at pose number 170, with its greatest distance from the distribution occurring at 175 before returning to the distribution. On the contrary to this, Learner 13 stays within distribution throughout the sequence, coming close to the upper bounds at point but remaining within the confidence region, indicating its variation is acceptable.

This visualisation demonstrates the pipelines ability to temporally determine where anomalies have occurred, and by how far they differ from the learnt distribution over natural variations. The latent signals from three of the native signers used to train the Motion Envelope for this sentence are visualised to demonstrate examples of the natural variation in SL between deaf individuals.

The decoded pose sequences for the two learners and one of the native signers are displayed below the plot, focusing on the region where the anomaly occurs. We take the latent signals between Pose Numbers 165 and 185 and decode them using the SkeletonVAE visualising every fifth pose within the range. At 165 the poses for the two learners are similar to each other, and slightly differ from the Native Signer, but stay within a margin of error. At 170, the arms of Learner 5 move in the opposite direction to Learner 13 and the native signer, who start to converge. At 175, Participant 13 is furthest in pose from the other two examples, with the wrong arm in the air. This is reflected as the point at which the participant is furthest from the learnt distribution. After this, the learner starts to move towards the direction of the learnt distribution, finally converging back with the other two examples as shown at pose number 185. This visualisation provides a spatial context of the error occurring in the skeleton space.

\section{Conclusion}
Sign Language Assessment tools are useful to aid in language learning and are underdeveloped. Previous work has focused on isolated signs, classification, or comparison against a single reference video to assess SL. In this paper, we proposed a novel assessment system to assess the comprehensibility of continuous SL sequences by modelling the natural distribution in human motion over multiple native deaf participants. 

Our experiments demonstrated that modelling using multiple native signers can lead to robust and interpretable results. This approach can be used to provide visual feedback to users in spatio-temporal contexts to aid in SL learning and assessment. We evaluated our results using real data from language learners and showed strong correlation between manually rated data and our approach.

As future work, we would like to expand our system to include non-manual feature assessment as these are important linguistic features that modify the meaning of SL.

\subsubsection{Acknowledgement.}
This work was supported by the SNSF project ‘SMILE II’ (CRSII5 193686), and the Innosuisse IICT Flagship (PFFS-21-47). This work reflects only the authors' views and the commission is not responsible for any use that may be made of the information it contains.

\bibliographystyle{splncs04}
\bibliography{main}
\end{document}